\documentclass[twocolumn]{article}

\usepackage{amsmath}
\usepackage{amsthm}
\usepackage{amsfonts}
\usepackage{booktabs}
\usepackage{rotating}
\usepackage{caption}
\usepackage{url}
\usepackage[scaled=.8]{beramono}

\providecommand{\e}[1]{\ensuremath{{\times}10^{#1}}}

\usepackage{authblk}
\usepackage{bm}
\newcommand{\bA}{{\bm A}}
\newcommand{\bH}{{\bm H}}

\newcommand{\bW}{{\bm W}}
\newcommand{\bI}{{\bm I}}
\newcommand{\bF}{{\bm F}}
\newcommand{\bE}{{\bm E}}
\newcommand{\bx}{{\bm x}}
\newcommand{\be}{{\bm e}}

\title{End-to-End Learning on Multimodal Knowledge Graphs}

\author[1]{W.X.~Wilcke\thanks{w.x.wilcke@vu.nl}}
\author[1]{P.~Bloem\thanks{p.bloem@vu.nl}}
\author[1]{V.~de~Boer\thanks{v.de.boer@vu.nl}}
\author[2]{R.H.~van~'t~Veer\thanks{rein.van.t.veer@geodan.nl}}
\author[1]{F.A.H.~van~Harmelen\thanks{frank.van.harmelen@vu.nl}}
\affil[1]{Department of Computer Science, Vrije Universiteit, Amsterdam, The Netherlands}
\affil[2]{Geodan, Amsterdam, The Netherlands}

\begin{document}

\maketitle

\begin{abstract}	
	Knowledge graphs enable data scientists to learn end-to-end on heterogeneous knowledge. However, most end-to-end
	models solely learn from the relational information encoded in graphs' structure:
	raw values, encoded as literal nodes, are either omitted completely or treated as regular nodes without
	consideration for their values. In either case we lose potentially relevant information which could have otherwise
	been exploited by our learning methods. We propose a multimodal message passing network which not
	only learns end-to-end from the structure of graphs, but also from their possibly divers set of multimodal node
	features. Our model uses dedicated (neural) encoders to naturally learn embeddings for node features belonging to
	five different types of modalities, including numbers, texts, dates, images and geometries, which are projected into
	a joint representation space together with their relational information. We implement and demonstrate our model on
	node classification and link prediction for artificial and real-worlds datasets, and evaluate the effect that each
	modality has on the overall performance in an inverse ablation study. Our results indicate that end-to-end
	multimodal learning from any arbitrary knowledge graph is indeed possible, and that including multimodal information
	can significantly affect performance, but that much depends on the characteristics of the data.

\end{abstract}

\section{Introduction}

\noindent The recent adoption of knowledge graphs by multinationals such as Google and Facebook has made them interesting targets
for various machine learning applications such as link prediction and node classification. Already, this interest has
lead to the development of message-passing models which enable data scientists to learn end-to-end\footnotemark from any
arbitrary graph. To do so, message-passing models propagate information over the edges of a graph, and can therefore be used to
exploit the relational information encoded in a graph's structure to guide the
learning process. The same approach has also been shown to work quite well on knowledge graphs, obtaining results that
are comparable to dedicated models such as \textit{RDF2Vec}~\cite{ristoski2016collection} and Weisfeiler-Lehman
kernels~\cite{shervashidze2011weisfeiler}. Nevertheless, by focusing on a single modality---the graphs' structure---we
are effectively throwing away a lot of other information that knowledge graphs tend to have, and which, if we were able
to include it in the learning process, has the potential of improving the overall performance of our models.

\footnotetext{In the context of this paper, we define ``end-to-end learning'' as the use of machine learning models
which operate directly on raw data, instead of relying on manually engineered features. In end-to-end learning, any
information in the data can, in principle, be used by the model. See \cite{wilcke2017the} for a more in-depth
discussion.}

\begin{figure*}[t]
  \includegraphics[width=\linewidth]{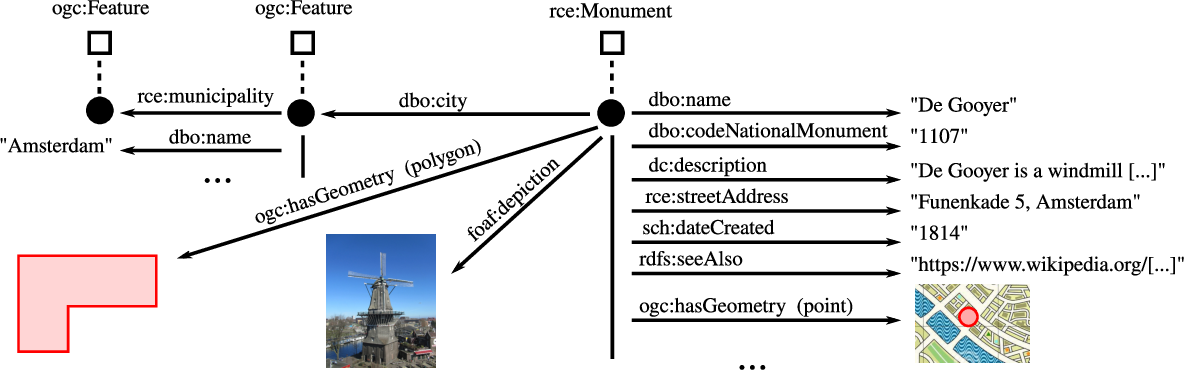}
  \caption{A simplified and incomplete example from the Dutch Monuments Graph showing a single monument with several
  attributes of different modalities.}
  \label{fig:examplegraph}
\end{figure*}

Combining information from multiple modalities is a topic that is already well studied for information stored in
\emph{relational} form (for instance in relational database management systems). Here too, we often encounter
\emph{heterogeneous} knowledge, containing information from a wide variety of modalities (such as language, audio, or
images). In \cite{wilcke2017the}, the case is made that to truly learn \emph{end-to-end} from a collection of
heterogeneous, multimodal data, we must design machine learning models that can consume these data in as raw a form as
possible, staying as close as we can to the original knowledge, and that we need to adopt a data model which can
represent our data in a suitable format, for which the knowledge graph is a natural choice. In other words, even when
our heterogeneous multimodal data is not initially represented as a knowledge graph, transforming it to this format is a
natural first step in an end-to-end multimodal machine learning pipeline. 

In this paper, we introduce and implement a multimodal message passing neural network, based on this principle, which
can directly consume heterogeneous multimodal data, represented as knowledge graph, and which itself can learn to
extract relevant information from each modality, based solely on the downstream task.


With the term \emph{knowledge graph} we mean any labeled multidigraph that is built on top of the \textit{Resource
Description Framework} (RDF). We consider the relational information of such a graph, encoded in its structure, as a
single modality. Other modalities that are commonly present in knowledge graphs are of numerical, textual, and temporal
nature, such as various measurements, names, and dates, respectively, and, to a lesser degree, of visual, auditory, and
spatial makeup. In a knowledge graph about monuments, for example, we might find that each monument has a detailed
description, a registration number, a year in which it was built, a few pictures from different angles, and a set of
coordinates (Figure~\ref{fig:examplegraph}). These and other attributes are encoded as raw values with corresponding
datatype annotations, called \textit{literals}, and tell us something about the objects they are connected to, called
\textit{entities}. However, most of this information is lost when we reduce the literals to identifiers, as is currently
common practice when we apply message passing networks to knowledge graphs. 

By reducing literals to identifiers, we discard any information that is contained in their contents, retaining only the
relational information encoded by their connections, and placing them on an equal footing with all other entities. This
means that we are effectively feeding our models a subset of the original and complete knowledge, but also that we are
depriving our models of the ability to compare inputs according to their modalities: measurements as numbers,
descriptions as language, coordinates as geometries, etc. As a result, our models are unable to distinguish between literals
that are closely together in the value space with those which are far apart. The name \textit{Mary}, for example, would
be seen as (dis)similar to \textit{Maria} as it would to \textit{Bigglesworth}, as would the integer value $\mathit{47}$
be to $\mathit{42}$ and $\mathit{6.626068 \times 10^{-34}}$. Instead however, we want our models to use this information
to guide their learning process. 

By enabling our models to naturally ingest literal values, and by treating these values according to their modalities,
tailoring their encodings to their specific characteristics, we stay much closer to the original and complete knowledge
that is available to us. We believe that doing so enables our models to create better internal representations of the
entities we are trying to learn over, potentially resulting in an increase in the overall performance of our models.  By
embedding this principle in the message passing framework, and by exploiting Semantic Web standards such as datatype annotations,
we embrace the idea that this enables us to learn end-to-end from any heterogeneous multimodal knowledge, as long as it
is represented as a knowledge graph.

In this work, we propose a multimodal message passing model which incorporates the information from a divers set of
multimodal node features. Our model uses dedicated vectorization strategies and (neural) encoders to naturally learn
embeddings for node	features belonging to five different types of modalities, including images and geometries, which are
projected into a joint representation space together with their relational information. We demonstrate our model on node
classification and link prediction for both artificial and real-worlds knowledge graphs, and evaluate the effect that each
modality has on the overall performance in an inverse ablation study. We also implement and publish our model as Python
package capable of learning from any arbitrary knowledge graph out of the box, exploiting Semantic Web standards to
automatically infer and incorporate multimodal information. 



To summarize, the main contributions of this paper are:
\begin{enumerate}
\item A machine learning model, embedded in the message passing framework, which can learn end-to-end from a
	heterogeneous knowledge, encoded as a knowledge graph, and which can naturally ingest literal values according to their modalities.  
\item An investigation of the potential usefulness of including information from multiple modalities, and the
	impact this has on the overall performance of our models. 
\item An implementation of our model (named the MR-GCN), which can learn from any arbitrary knowledge graph, and which
	exploits Semantic-Web standards to automatically infer and incorporate multimodal information.
\end{enumerate}

\noindent Our intent is emphatically \emph{not} to show that our implementation achieves any kind of state-of-the-art, or even
to measure its performance against related models. Rather, we aim to demonstrate that
1) by including as much of the original knowledge as possible, in as natural of a fashion as possible, we can, in
certain cases, help our models obtain a better overall performance, and that 2) a model can be trained end-to-end on a
heterogeneous knowledge graph such that it learns purely from the downstream task which patterns to extract from each
modality.

\section{Related Work}

\noindent Machine learning from multimodal sources is a well-studied problem. A good introduction to this problem and its many
perspectives is given by \cite{baltruvsaitis2018multimodal}. According to their taxonomy, our approach is one of
\emph{late fusion} by first encoding modalities using dedicated neural encoders, after which the resulting encodings
are projected in a joint representation space. Different from most other research in this field we are not
interested in \textit{translation} (mapping one modality to another) nor in \textit{alignment} (aligning the same
subject over multiple modalities). Rather, information in a given modality is only ever used to learn node embeddings
with the intent to improve the learning process by including as much of the original knowledge as possible.

\subsection{Knowledge Graph Embeddings}

\noindent Graph embedding techniques aim to represent graphs in a lower-dimensional space, making them more suitable
to learn over. Numerous embedding techniques have been proposed over the years, and typically differ in which operations
they apply between the node and edge embeddings, and which scoring function they use. Popular methods are those based on
matrix factorization, random walks, translation models, and, more recently, deep neural
networks~\cite{cai_comprehensive_2018}. Our approach falls in the latter group of methods, for its use of a
message-passing network. A thorough overview of the different embedding methods can be found in one of the many recent
survey papers, for example~\cite{cai_comprehensive_2018} and~\cite{wang2017knowledge}. Here, we will limit ourselves to
the graph embedding methods that consider multimodal information.


Various approaches have explored using information from one or more additional modalities in machine learning models for
knowledge graphs. In most cases, only a singly additional modality is included, always of numerical, textual, or visual
nature~\cite{gesese2019survey}. This differs from our method, which also supports temporal and spatial literals. Our
methods also differs from most other approaches in that we address how information from different modalities can be
1) extracted from a graph, and 2) vectorized with minimal loss of information. 

An early work described in~\cite{nickel2012factorizing} proposes an extension to the RESCAL~\cite{nickel2011three} tensor factorization
method which can also cope with textual attributes. This is done by introducing an additional tensor which is
factorized together with the tensor holding the relational information. A similar separation is proposed
by~\cite{cochez2018first}, who generate a separate co-occurrence matrix for the relational and textual information,
and which are then summed to produce the final embeddings. Both these methods scale well due to their use of basic matrix
operations, whereas scalability remains a challenge for many message-passing models such as the one used in our approach. 

In~\cite{kristiadi2018incorporating}, the authors introduce a learnable function, called \textit{LiteralE}, which
replaces every entity embedding by a new embedding that is the fusion of the original entity embedding and its direct
numerical attributes. The resulting vector representation can then be used in an arbitrary translation-based model.
The fusion step is similar to our approach in that the embeddings of neighbouring nodes coalesce into the target entity,
except that our model does this for every node (entity or literal), up to an arbitrary depth (determined by the
number of layers in the message-passing network), and only after the modalities have been encoded according to
their specific characteristics. 

The authors of~\cite{de2018towards} propose an extension to LiteralE that incorporates textual features which they
generate by performing entity resolution on (part of) the identifiers of entities and relations. The results are then 
mapped to integers and passed to LiteralE together with the corresponding entities. 

A slightly different approach is proposed by~\cite{wu2018knowledge}, who perform a joint optimization of an existing
translation model (\textit{TransE}~\cite{bordes2013translating}) and a regression model specifically designed by the authors for numerical
features. The work in~\cite{xie2016representation} uses a similar approach, but for textual rather than numerical 
attributes and with a self-defined translation model instead of a regression model. Similar to our work, the authors use
a CNN as encoder for textual attributes, but where our model employs a temporal CNN with one-hot encoded text as input,
the authors here use a language-agnostic CNN with pretrained \textit{word2vec}~\cite{mikolov2013distributed} embeddings as input.

Another extension to an arbitrary translation model is proposed in~\cite{xie2016image}, who use a proven CNN
architecture to learn image embeddings, which are then used in a self-defined translation model.
For entities with more than one image attribute, the images embeddings are merged into one final embedding which is kept
separate from the entity embedding to which they belong. Our model differs in that all neighbouring nodes, and not just
images, coalesce into the corresponding entity embedding: separate image embeddings only exist prior to fusion.

Different from translation-based approaches is the work in~\cite{tay2017multi}, who propose using a dual network 
architecture with a binary classifier to learn relational information and a regression model to learn numerical
information. A joint optimization is used to train the model.

More modalities are considered by~\cite{emnlp18}, who incorporate numerical and textual literals, as well as images. The
numerical features are encoded using a feed-forward layer, which projects the values to a higher-dimensional space. For
short strings, the authors employ a character-based GRU, whereas a language-aware CNN is used in combination with word
sequences for longer strings. Finally, for images, the authors use the last hidden layer of a pretrained network on
ImageNet~\cite{deng2009imagenet}. The resulting embeddings are then paired with their corresponding entity embeddings (generated using a
feed-forward network) and ultimately scored using DistMult. The use of dedicated neural encoders per modality is similar
to our work, except for numerical features, which we feed directly to the message-passing network after normalization.
Also similar is the use of different encoders for text of different lengths, but rather than have completely different
models and input requirements, we employ three temporal CNNs of increasing size for short, medium, and long strings.

All the reviewed models are simple embedding models, based on basic matrix operations or on a score function applied to
triples. By contrast, our approach includes a message passing layer, allowing multimodal information to be propagated
through the graph, several hops and from \emph{all} (direct and indirect) neighbours.


\section{Preliminaries}

\noindent Knowledge graphs and message passing neural networks are integral components of our research. We will here briefly introduce both
concepts.

\subsection{Knowledge Graphs} 

\noindent For the purposes of this paper we define a \textit{knowledge graph} $G = (\mathcal{V},
\mathcal{E})$ over modalities $1, \ldots, \mathcal{M}$ as a labeled multidigraph defined by a set of nodes $\mathcal{V}
= \mathcal{I} \cup \bigcup\{\mathcal{L}^m | m \in \mathcal{M}\}$ and a set of directed edges $\mathcal{E}$, and with $n
= |\mathcal{V}|$. Nodes belong to one of two categories: entities $\mathcal{I}$, which represent objects (monuments,
people, concepts, etc.), and literals $\mathcal{L}^m$, which represent raw values in modality $m \in \mathcal{M}$
(numbers, strings, coordinates, etc.). We also define a set of relations $\mathcal{R}$, which
contains the edge types that make up $\mathcal{E}$. Relations are also called \textit{predicates}.

Information in $G$ is encoded as triples $\mathcal{T}$ of the form $(h, r, t)$, with head $h \in \mathcal{I}$, relation
$r \in \mathcal{R}$, and tail $t \in \mathcal{I} \cup \mathcal{L}^1 \cup \ldots \cup \mathcal{L}^m$. The combination of
relations and literals are also called \textit{attributes} or \textit{node features}.

See Figure~\ref{fig:examplegraph} for an example of knowledge graph with seven nodes, two of which are entities and the
rest literals. All knowledge graphs in this paper are
stored in the \textit{Resource Description Framework} (RDF) format~\cite{lassila1998resource}, but our model can be applied to
any graph fitting the above definition.

\subsection{Message Passing Neural Networks}
\label{ssc:mpnn}

\noindent A \emph{message passing neural network} \cite{gilmer2017neural} is a graph neural network
model that uses trainable functions to propagate node embeddings over the edges of the neural network. One simple
approach to message passing is the graph convolutional neural network (GCN)~\cite{kipf2016semi}. The
R-GCN~\cite{schlichtkrull2018modeling}, on which we build, is a straightforward extension to the knowledge graph setting.

Let $\bH^0$ be a $n \times q$ matrix of $q$ dimensional node embeddings for all $n$ nodes in the graph. That is, the
$i$-th row of $\bH^0$ is an embedding for the $i$-th node in the graph\footnotemark, The R-GCN computes an updated $n
\times l$ matrix $\bH^1$ of $l$-dimensional node embeddings by the following computation (the \emph{graph convolution}):
\begin{equation}
	\label{eq:rgcn}
 \bH^1 = \sigma\left(\sum_{r\in \mathcal{R}} \bA^r\bH^0\bW^r \right)
 \end{equation}
Here, $\sigma$ is an activation function like
ReLU, applied element-wise. $\bA^r$ is the row-normalised adjacency matrix for the relation $r$ and $\bW^r$ is a $q
\times l$ matrix of learnable weights. This operation arrives at a new node embedding for a node by averaging the
embeddings of all its neighbours, and linearly projecting to $l$ dimensions by $\bW^r$. The embeddings are then summed
over all relations and a non-linearity $\sigma$ is applied. 

\footnotetext{The standard R-GCN does not distinguish between literals and entities. Also, literals with the same value
are collapsed into one node, therefore $n \leq |\mathcal{V}|$.}

To allow information to propagate in both directions along an edge, all inverse relations are added to the predicate
set. The identity relation is also added (for which $\bA^r = \bI$) so that the information in the current embedding can,
in principle, be retained. To reduce overfitting, the weights $\bW^r$ can be derived from a smaller set of \emph{basis
weights} by linear combinations (see the original paper for details).

To use R-GCNs for entity classification with $c$ classes, the standard approach is to start with one-hot vectors as
initial node embeddings (that is, $\bH^0 = \bI$). These are transformed to $h$-dimensional node embeddings by a first
R-GCN layer, which are transformed to $c$-dimensional node embeddings by a second R-GCN layer. The
second layer has a row-wise softmax non-linearity, so that the final node embeddings can be read as class probabilities.
The network is then trained by computing the cross-entropy loss for the known labels and backpropagating to update the
weights. Using more than two layers of message passing does not commonly improve performance with current message
passing models.

For link prediction, the R-GCNs can be viewed as encoder in a graph auto-encoder. In that role, the R-GCNs learn
node embeddings that are used by a decoder to reconstruct the edges in the graph. As before, the standard approach
for the R-GCNs is to have one or two layers, and to start with one-hot vectors as initial node embeddings.
However, because we are now interested in the node embeddings themselves, the softmax on the end
is replaced with an activation function like ReLU, applied element-wise. The decoder consists of a triple scoring
function $s: \mathcal{V} \times \mathcal{R} \times \mathcal{V} \mapsto \mathbb{R}$, for which ideally holds that 
$s(h, r, t) > s(x, y, z)$ if $(h, r, t)$ exists and $(x, y, z)$ does not. 

In this work, we use DistMult~\cite{yang_embedding_2014} for our decoder, which is known to perform well
on link prediction tasks while keeping the number of parameters low~\cite{ruffinelli_you_2019}. DistMult uses the 
following bilinear scoring function:
\begin{equation}
	\label{eq:distmult}
	s(\mathbf{y}_{v_i}, r, \mathbf{y}_{v_j}) = \mathbf{y}^{T}_{v_i} \mathrm{diag}(\mathbf{R}_r) \mathbf{y}_{v_j}
\end{equation}
Here, $\mathbf{y}_{v_i}$ and $\mathbf{y}_{v_j}$ are the output of the encoder for nodes $v_i, v_j \in \mathcal{V}$, and
$\mathbf{R}_r$ the embedding belonging to relation $r \in \mathcal{R}$. Both encoder and decoder are trained by
minimizing the binary-cross entropy loss\footnotemark over the output of Equation~\ref{eq:distmult} for both positive
and negative samples (negative sampling)~\cite{schlichtkrull2018modeling}. The set of negative samples $\mathcal{T}^-$
can be obtained by randomly corrupting the head or tail of a portion ($\frac{1}{5}$) of the triples in
$\mathcal{T}$.

\footnotetext{A margin ranking loss is used in the original DistMult paper.}

\section{A Multimodal Message Passing Network}

\begin{figure}[t]
  \includegraphics[width=\linewidth]{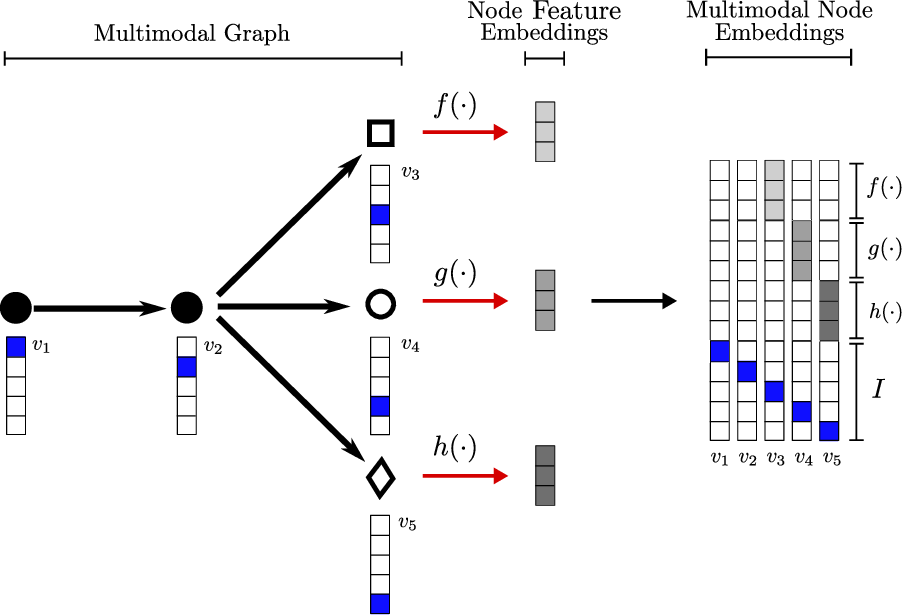}
  \caption{Overview of how our model creates multimodal node embeddings for nodes $v_1$ to $v_5$. Solid circles
	  represent entities, whereas open shapes represent literals of different modalities. The nodes' feature
	  embeddings are learned using dedicated (neural) encoders (here $f$, $g$, and $h$), and concatenated to their
  identity vectors $I$ to form multimodal node embeddings, which are fed to a message passing network.}
  \label{fig:model}
\end{figure}

\noindent We introduce our model as an extension to message passing networks which can learn end-to-end from the structure of an
arbitrary graph, and for which holds that $\bH^0=\bI$. To do so, we let $f(\cdot)$, $g(\cdot)$, and $h(\cdot)$ be feature encoders that output feature
embeddings of lengths $\ell_f$, $\ell_g$, and $\ell_h$ for nodes $v_i \in \mathcal{V}$. We define $\bF$ as the $n
\times f$ matrix of multimodal feature embeddings with $f = \ell_f + \ell_g + \ell_h$, and concatenate $\bF$ to the
identity matrix $\bI$ to form multimodal \emph{node} embeddings:
\begin{equation}
	\label{eq:input}
	\bH^0 = [\bI~\bF]
\end{equation}
\noindent of size $n \times q$ (Fig.~\ref{fig:model}).

Embedding matrix $\bH^0$ is fed together with $\bA^r$ to a message passing network, such as the R-GCN. Both encoders and
network are trained end-to-end in unison by backpropagating the error signal from the network through the encoders all
the way to the input. 

\subsection{Modality Encoders}
\label{sec:modalities}

\noindent We consider five different modalities which are commonly found in knowledge graphs. We forgo discussing
relational information---the sixth modality---as that is already extensively discussed in related work on message
passing networks. For numerical information, we use a straightforward one-to-one encoding and let the message-passing
layers handle it further. For all other modalities we use neural encoders: a feed-forward neural network for temporal
information, and convolutional neural networks (CNN) for textual, spatial, and visual information. Each of these will be
discussed next. We will also discuss the preceding vectorization process, which, if done poorly, can results in a loss
of information.

In the following, we let $\be^m_i$ be the embedding vector of node $v_i$ for modality $m$. The concatenation of
a node's identity vector and all its feature embedding vectors $\be^m_i$ for every $m \in \mathcal{M}$ equals the $i$-th
row of $\bH^0$.

\subsubsection{Numerical Information} 

\noindent Numerical information encompasses the set of real numbers $\mathbb{R}$, and corresponds to literal values with
a datatype declaration of \texttt{XSD:double}, \texttt{XSD:float}, and \texttt{XSD:decimal} and any subtype thereof. For
these, we can simply take the normalized values as their embeddings, and feed these directly to the message-passing
layers. We also include values of the type \texttt{XSD:boolean} into this category, but separate their representations 
from those of real numbers to convey a difference in semantics.

More concretely, for all nodes $v_i \in \mathcal{V}$ holds that $\be^{num}_i$ is the concatenation of their numerical
and boolean components, encoded by functions $f_{num}$ and $f_{bool}$, respectively. Here, $f_{num}(v_i) = v_i$ if $v_i$ is
a literal node with a value in $\mathbb{R}$. If $v_i$ is a boolean instead, we let $f_{bool}(v_i)$ be $1.0$ if $v_i$ is
\texttt{true} and $-1.0$ if $v_i$ is \texttt{false}. In both cases, we represent missing or erroneous values with $0.0$
(we assume a normalization between -1 and 1).

\subsubsection{Temporal Information}

\noindent Literal values with datatypes which follow the \textit{Seven-property model}\footnotemark such as \texttt{XSD:time},
\texttt{XSD:date} and\\\texttt{XSD:gMonth}, are treated as temporal information.
Different from numerical values, temporal values contain elements that are defined in a circular value space and which
should be treated as such. For example, it is inaccurate to treat the months December and January as if they were 11 months apart,
as would be implied by directly feeding the months' number to our models. Instead, we can represent this as
\begin{equation}
	\label{eq:circ}
	f_{trig}(\phi, \psi) = [sin(\frac{2\pi \phi}{\psi}), cos(\frac{2\pi \phi}{\psi})]
\end{equation}
with $\psi$ the number of elements in the value space (here 12), $\phi$ the integer representation of the element
we want to encode, and $f_{trig}$ a trigonometric function in our encoder. This ensures that the representation of January
is closer to that of December than it is to that of March.

\footnotetext{https://www.w3.org/TR/xmlschema11-2}

We can use this representation for all other circular elements, such as hours ($\psi=24$) and decades ($\psi=10$). When dealing with
years however, we represent smaller changes more granular than larger changes: years are split into
centuries, decades, and (single) years fragments, with decades and years treated as circular elements but with centuries
as numerical values (we limit our domain to years between $-9999$ and $9999$). 

Once vectorized, the vector representation $\mathbf{v}_i$ is fed to a feed-forward neural network $f_{temp}$ with input and output dimensions
$n_{in}$ and $n_{out}$, respectively, and for which holds that $n_{in} < n_{out}$, such that $\be^{temp}_i = f_{temp}(\mathbf{v}_i)$.

\subsubsection{Textual Information}

\noindent Vector representations for textual attributes with the datatype \texttt{XSD:string}, or any subtype thereof,
and \texttt{XSD:anyURI} are created using a character-level encoding, as proposed in \cite{zhang2015character}. For this
purpose, we let $\bE^s$ be a $|\Omega| \times |s|$
matrix representing string $s$ using vocabulary $\Omega$, such that $\bE^s_{ij}=1.0$ if $s_j = \Omega_i$, and $0.0$
otherwise.

A character-level representation enables
our models to be language agnostic and independent of controlled vocabularies (allowing it to cope with colloquialisms and
identifiers for example), as well as provide some robustness to spelling errors. It also enables us to forgo the
otherwise necessary stemming and lemmatization steps, which would remove information from the original text.
The resulting embeddings are optimized by running them through a temporal CNN $f_{char}$ with output dimension $c$, such that 
$\be^{textual}_i = f_{char}(\bE^{v_i})$ for every node $v_i$ with a textual value.

\subsubsection{Visual Information}

\noindent Images and other kinds of visual information (e.g.\ videos, which can be split in frames) can be included in a knowledge
graph by either linking to them or by expressing them as binary string literals\footnotemark which are incorporated in the graph itself 
(as opposed to storing them elsewhere). In either
case, we first have to obtain the raw image files by downloading and/or converting them.

\footnotetext{In~\cite{peter2020kgbench}, we advocate the use of KGBench's \texttt{base64Image}
for this purpose.}

Let $im_i$ be the raw image file as linked to or encoded by node $v_i$. We can represent this image as a tensor
$\bE^{im_i}$ of size $channels \times width \times height$, which we can feed to a two-dimensional CNN $f_{im}$ with
output dimension $c$, such that $\be^{visual}_i = f_{im}(\bE^{im_i})$ for the image associated with 
node $v_i$.

\subsubsection{Spatial Information}

\noindent Spatial information includes points, polygons, and any other spatial features that consist of one or more coordinates.
These features can represent anything from real-life locations or areas to molecules or more abstract mathematical
shapes. Literals with this type of information are commonly expressed using the \textit{well-known text representation}
(WKT) and carry the \texttt{OGC:wktLiteral} datatype declaration. The most elementary spatial feature is a
coordinate (point geometry) in a $d$-dimensional space, expressed as \texttt{POINT}($x_1 \ldots x_d$), which can be
combined to form more complex types such as lines and polygons. 

We can use the vector representations proposed in \cite{van2018deep} to represent spatial features.
Let $\bE^{sf}$ be the $|\bx| \times |sf|$ matrix representation for spatial feature $sf$
consisting of $|sf|$ coordinates, and with $\bx$ the vector representation of one such coordinate. Vector $\bx$ holds
all of the coordinate's $d$ points, followed by its other information (e.g.\ whether it is part of a polygon) 
encoded as binary values. For spatial features with more than one coordinate, we also need to separate their location from
their shape to ensure that we capture both these components. To do so, we encode the location in $\mathbb{R}^d$ by taking 
the mean of all coordinates that makeup the feature. To capture the shape, we compute the global mean of all spatial
features in the graph, and subtract this from their coordinates to place their centre around the origin. 

We feed the vector representations using a temporal CNN $f_{sf}$ with output dimension $c$, such that 
$\be^{spatial}_i  = f_{sf}(\bE^{v_i})$ for all nodes $v_i$ which express spatial features.

\section{Implementation}

\noindent We implement our model using the R-GCN as our main building block, onto which we stack our various
encoders. We call this a multimodal R-GCN (MR-GCN). The R-GCN is a suitable choice for this purpose, as it can learn
end-to-end on the structure of relational graphs, taking relation types into account. Our implementation is available
as Python package\footnotemark, and can be used with any arbitrary knowledge graph in RDF format. 

\footnotetext{Code available at https://gitlab.com/wxwilcke/mrgcn}

In the simplest case, when we are only interested in learning from the graph's structure or when no multimodal
information is present in the graph, we let the initial node embedding matrix $\bH^0$ be the nodes' $n \times n$
identity matrix $\bI$ (i.e.\ $\bH^0=\bI$). This reduces the MR-GCN to a plain R-GCN. To also include multimodal
information in the learning process, we let $\bF$ be the $n \times f$ feature embedding matrix instead and concatenate
this to $\bH^0$ as in Equation~\ref{eq:input} to form $\bH^0=[\bI~\bF]$. 

To accurately determine the most suitable encoder for each encountered literal, the MR-GCN exploits Semantic-Web
standards to automatically infer this from the graph's datatype annotations. Supported datatypes include many XSD
classes, such as numbers, strings, and dates, as well as OGC's \texttt{wktLiteral} for spatial information, and
KGbench's \texttt{base64Image} for binary-encoded images~\cite{peter2020kgbench}. These modalities are assumed to be
encoded directly in the graph, as opposed to reading them from separate files.

To cope with the increased complexity brought on by including node features we optimized the MR-GCN
for sparse matrix operations by splitting up the computation of Equation~\ref{eq:rgcn} into the sum of the structural
and feature component. For this, we once again split $\bH^0$ into identity matrix $\bH_I=\bI$ and feature matrix
$\bH_F^0=\bF$, and rewrite the computation as 

\begin{equation}
	\label{eq:mrgcn}
    \bH^1 = \sigma\left(\sum_{r\in \mathcal{R}} \bA^r\bH_I\bW_I^r + \bA^r\bH_F^0\bW_F^r \right)
\end{equation}

\begin{figure}[t]
  \includegraphics[width=\linewidth]{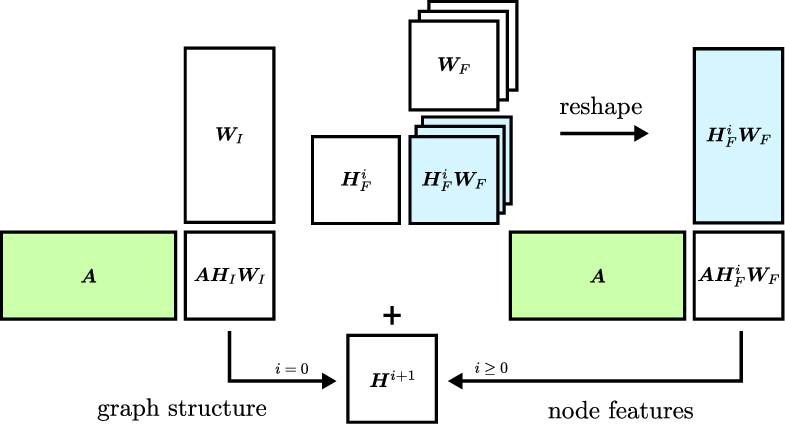}
  \caption{Graphical depiction of our implementation of Equation~\ref{eq:mrgcn}, shown as matrix operations. The
	  output of layer $i$, $\bH^{i+1}$, is computed by summing the structure and node feature
  components. If $i>0$, then $\bH^i_F = \bH^i$ and $\bA\bH_I\bW_I = 0$.}
  \label{fig:implementation}
\end{figure}

Here, $\bW_I^r$ and $\bW_F^r$ are the learnable weights for the structural and feature components, respectively. For
layers $i>0$ holds that $\bH^i_F = \bH^i$, and that $\bA^r\bH_I\bW_I^r = 0$. Note that because $\bA^r\bH_I = \bA^r$, we
can omit this calculation when computing Equation~\ref{eq:mrgcn}, and thus also no longer need $\bH_I$ as input.
Figure~\ref{fig:implementation} illustrates this computation as matrix operations.

To support link prediction, the MR-GCN implements the DistMult~\cite{yang_embedding_2014} bilinear scoring function, shown
in Equation~\ref{eq:distmult}. To reduce the number of parameters, we simulate relation embeddings $diag(\mathbf{R})$ by
a $|\mathcal{R}| \times h$ matrix, with each row representing the diagonal of a theoretical relation embedding
$\mathbf{R}_r$.

\subsection{Neural Encoders}

\noindent The MR-GCN implements neural encoders for all modalities listed in Section~\ref{sec:modalities}. For temporal
information, we use a single layer fully connected feed-forward neural network of which the dimensions depend on the
datatype, as shown in Table~\ref{tab:encodertemp}. The three other neural encoders are all implemented using CNNs, each
initiated using $\mathcal{N}(0, 1)$ and with an output dimension of 128.

For our visual encoder, we use the efficient MobileNet architecture from~\cite{howard2017mobilenets}, which provides a
good performance with relatively few parameters. For spatial information, we use a temporal CNN similar to that used
in~\cite{van2018deep}, which has 3 convolutional layers, each followed by $\mathrm{ReLU}$, and 3 dense layers
(Table~\ref{tab:encodergeom}). A similar setup is used for textual information, except that we use different
architectures for short ($\ell < 20$), medium ($20 < \ell < 50$), and long ($\ell > 50$) strings, with $\ell$ denoting
their length. The architecture for medium-length strings is listed in Table~\ref{tab:encodertext}, whereas for long
strings we double the number of filters to 128 and let the first dense layer have 1024 hidden nodes. For short strings,
we omit the last convolutional and dense layer (layer 4 and 7), and reduce the number of hidden nodes in the first dense
layer to 256.

\begin{table}[t]
	\caption{Configurations of the neural encoder for temporal information with $h$ hidden nodes and output dimension
		$n_{out}$, listed per tested datatype. Note that $n_{in} = h$}
\label{tab:encodertemp}
\centering
\begin{tabular}{lcc}
\toprule
Datatype & $h$ & $n_{out}$\\\midrule
\texttt{XSD:gYear} & 6 & 2 \\
\texttt{XSD:date} & 10 & 4 \\
\texttt{XSD:dateTime}& 14 & 6\\\bottomrule
\end{tabular}
\end{table}

\begin{table}[t]
	\caption{Configuration of the textual encoder for medium-length strings with 4 convolutional layers (top) and 3 dense layers (bottom). For
		pooling layers, \textit{max(k/s)} lists kernel size (\textit{k}) and stride (\textit{s}), or
		\textit{max}$(\cdot)$ when it depends on the input sequence length.}
\label{tab:encodertext}
\centering
\begin{tabular}{ccccc}
\toprule
Layer  & Filters & Kernel & Padding & Pool\\\midrule
1 & 64 & 7 & 3 & max(2/2)\\
2 & 64 & 7 & 3 & max(2/2)\\
3 & 64 & 7 & 3 & -\\
4 & 64 & 7 & 2 & max($\cdot$) \\\bottomrule
\end{tabular}

\bigskip

\begin{tabular}{cc}
	\toprule
	Layer  & Dimensions\\\midrule
	5 & 512\\
	6 & 128\\
	7 & 128\\\bottomrule
\end{tabular}
\end{table}

\begin{table}[t] \caption{Configuration of the spatial encoder with 3 convolutional layers (top) and 3
	dense layers (bottom). For pooling layers, \textit{max(k/s)} lists kernel size (\textit{k}) and stride (\textit{s}),
whereas \textit{avg}$(\cdot)$ depends on the input sequence length.}
\label{tab:encodergeom}
\centering
\begin{tabular}{ccccc}
\toprule
layer  & filters & kernel & padding & pool\\\midrule
1 & 16 & 5 & 2 & max(3/3)\\
2 & 32 & 5 & 2 & -\\
3 & 64 & 5 & 2 & avg($\cdot$) \\\bottomrule
\end{tabular}

\bigskip

\begin{tabular}{cc}
	\toprule
	layer  & dimensions\\\midrule
	4 & 512\\
	5 & 128\\
	6 & 128\\\bottomrule
\end{tabular}
\end{table}

The output of layer $i$ from all encoders for all nodes in $\mathcal{V}$ are concatenated to form $\bH_F^i$, which is passed to
Equation~\ref{eq:mrgcn} together with $\bA^r$. 

\section{Experiments}

\noindent We evaluate the MR-GCN on node classification and link prediction while varying the modalities which are
included in the learning process\footnotemark. For this purpose, we compute the performance for each combination of structure and modality,
as well as all modalities combined, and evaluate this against using only the relational information. To eliminate any
confounding factors in real-world knowledge that might influence the results, we will first evaluate the MR-GCN on synthetic
knowledge (Section~\ref{sec:evalsynth}) before testing our implementation on real-world datasets
(Section~\ref{sec:evalrw}).
\footnotetext{Datasets available at \url{https://gitlab.com/wxwilcke/mmkg}}

Another dimension that we vary is how much raw information is already implicitly encoded in the structure of a graph by
having literals nodes with an indegree greater than one. This occurs when literals with the same value are coalesced into
a single node, and is the standard approach to represent knowledge graphs in graph form. Encoding this
information in a graph's structure influences the potential gain in performance we can obtain by including node 
features in the learning process, possibly even masking it. Consider, for example, a classification problem in which
a small range of literals perfectly separates our classes: when this information is already encoded in the structure there might
be little to gain by enabling our models to compare these literals by their values, whereas doing so if this information
is \textit{not} encoded in the structure might yield a significant performance boost. In our experiments, we will use
the term \textit{split literals} to refer to the representation that keeps literals with the same value as separate
nodes (i.e.\ indegree $=$ 1), and use the term \textit{merged literals} to refer to alternative representation in which
literals with the same value are coalesced (i.e.\ indegree $\ge$ 1).

For our node classification experiments we use an architecture similar to the plain R-GCN (Section~\ref{ssc:mpnn}).
Concretely, we employ a two-layered MR-GCN with 32 hidden nodes, and with an element-wise ReLU activation function 
after the first layer. A row-wise softmax non-linearity is added to the second layer to output class probabilities. The
network is trained by minimizing the cross-entropy loss in full batch mode with Adam for 400 epochs with an initial
learning rate of $0.01$. 

For each configuration we report the mean classification accuracy and 95\% confidence interval over 10 runs. To check 
the results on statistical significance, we use the Stuart-Maxwell marginal homogeneity test which
tests whether two multi-class models have the same distribution of predictions~\cite{maxwell_comparing_1970,stuart_test_1955}.
To obtain a single set of predictions per configuration for this purpose, we use a majority vote amongst the ordered output
from all 10 runs. 

Our link prediction experiments likewise use a graph auto-decoder architecture similar to the plain R-GCN
(Section~\ref{ssc:mpnn}). More specific, we employ a single-layered MR-GCN with 200 hidden nodes, with an element-wise
ReLU activation function at the end, and with DistMult as triple scoring function. We train the network by minimizing
the binary cross-entropy loss in full batch mode with Adam for 1000 epochs with an initial learning rate of $0.01$. 

For each configuration we report the filtered mean reciprocal rank (MRR) and hits@$k$ with $k \in \{1,3,10\}$ over 5 runs, as
well as the 95\% confidence interval and statistical significance computed over the MRR\footnotemark. To check for
statistical significance, we use the computational-intensive randomised paired t-test~\cite{cohen_empirical_1995},
as suggested by~\cite{yeh_more_2000}, which tests whether two ordered sets of ranks have the same distribution of mean
differences. Note that, with this method, the minimal achievable p-value depends on the size of the test set. As with
classification, we obtain a single set of ranks per configuration by majority vote.

\footnotetext{As the hits@$k$ is derived from the MRR, no new information is gained by also computing the confidence
interval and statistical significance of the former.}

\subsection{Evaluation on Synthetic Knowledge}
\label{sec:evalsynth}

\begin{figure}[t]
  \includegraphics[width=\linewidth]{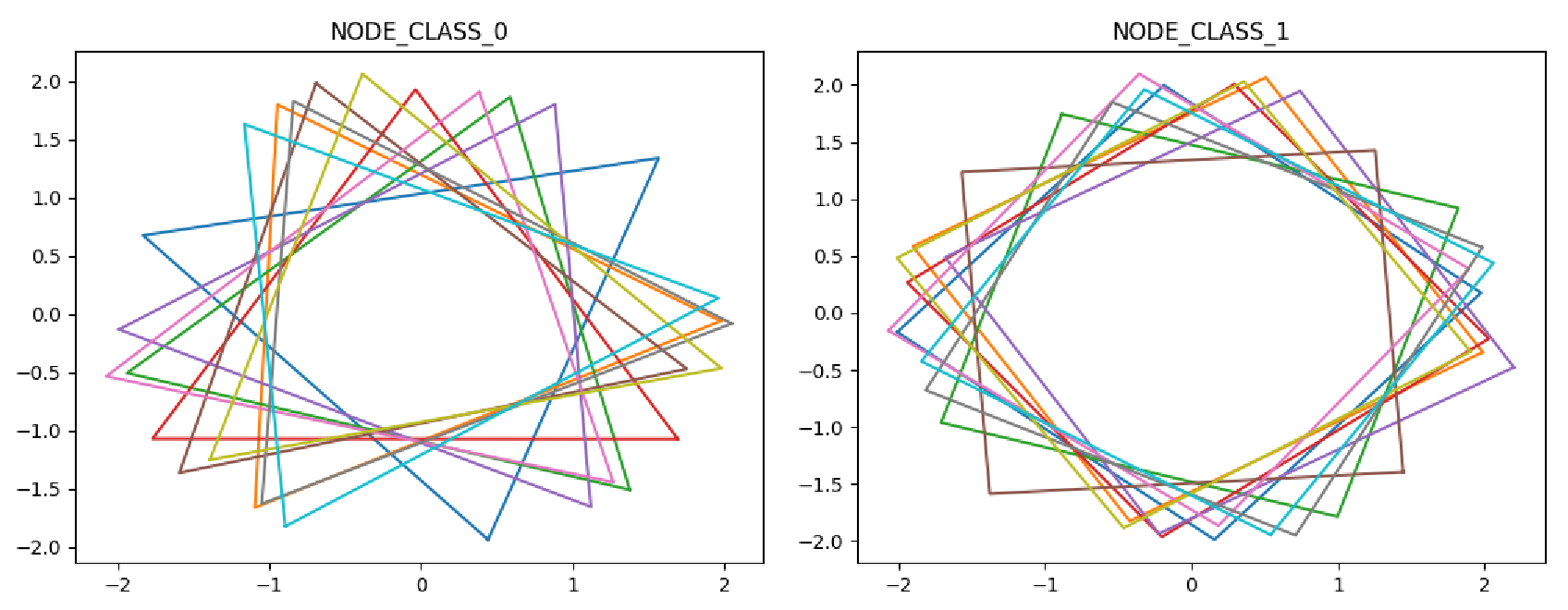}
  \caption{Geometries belonging to 10 randomly-sampled entities per class from the SYNTH dataset. Apart from the number
  of points (which our model is agnostic to) the only difference between classes is the shape.}
  \label{fig:geom}
\end{figure}
\begin{figure}[t]
  \includegraphics[width=\linewidth]{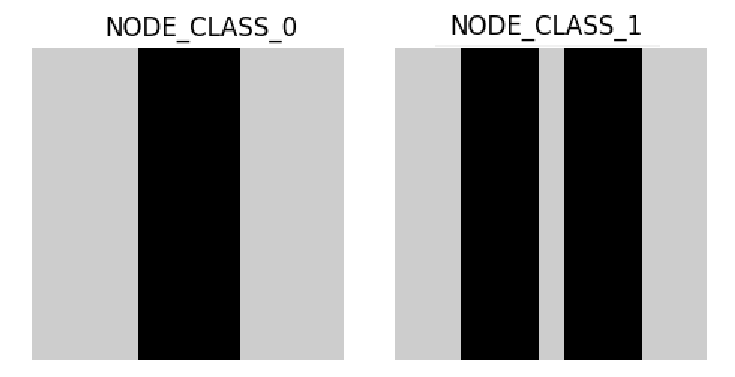}
  \caption{Images belonging to entities per class from the SYNTH dataset, shown here without the noise normally present
	  to ensure different string representations with a class.}
  \label{fig:im}
\end{figure}

\noindent We first evaluate the performance of the MR-GCN on synthetic data. These data serve as a controlled environment which
enables us to eliminate any confounding factors in real-world data that would otherwise influence the results, ensuring
that any observed difference can be confidently attributed to the addition or removal of a certain modality. For this
purpose, we generated\footnotemark a synthetic knowledge graph (SYNTH) that contains strong multimodal signals, but
which lacks relational information. General and modality-specific statistics are listed in Table~\ref{tab:datasets}
and~\ref{tab:datatypes}, respectively. 

\footnotetext{Code available a \url{https://gitlab.com/wxwilcke/graphsynth}}

The SYNTH dataset consists of 16,384 entities, all labeled, from two distinctly different classes, and connected by a
random graph structure that is generated using the Watts–Strogatz algorithm. Each entity is provided with literals of
different datatypes, encompassing all five modalities listed in Section~\ref{sec:modalities}. To ensure that the
learning problem is both manageable and challenging, the literal values were drawn from two narrow and slightly overlapping
distributions, with noise added where necessary. These distributions were generated with the corresponding modality in
mind: numbers and years where drawn from Gaussian distributions, dates and times were sampled around specific months and
hours, respectively, and strings were generated by combining a class-specific keyword with randomly sampled words from a
dictionary. This principle is also shown in Figure~\ref{fig:geom} for geometries, which only differ in shape\footnotemark to force
our model to capture this characteristic. Similarly in Figure~\ref{fig:im} for images, which are unique per class and
to an extent robust to transformations (e.g., scale, rotation, translation).

\footnotetext{The neural encoders in our model are agnostic to the number of points.}

\subsubsection{Node Classification Results}

\noindent Table~\ref{tab:resultssynthclassification} reports the mean classification accuracy over 10 runs on SYNTH, together with
its 95\% confidence interval and corresponding p-values. We use \textit{value\_merged} [\textit{value\_split}] to express
the performances in the merged and split configurations, respectively.

Overall, the results indicate that, for all modalities and literal
configurations, including node features considerably increases the performance over that of the baseline (structure
only). When all node features are taken into account, this performance increase raises the accuracy from near random
(0.616 [0.495]) to near perfect (0.995 [0.996]). All reported performance gains are statistically significant, with as
highest p-value 5.21\e{-04}.

When comparing the performance gain per modality it is evident that this differs widely between modalities: including
just textual or spatial information increases the performance to a near perfect accuracy of 0.995 [0.996] and 0.957
[0.949], respectively, whereas including only visual information just provides a slight (although still significant)
gain to an accuracy of 0.642 [0.556]. The remaining two modalities---numerical and temporal information---lie in between
these two extremes and provide a moderate performance boost with an accuracy of 0.744 [0.785] and 0.763 [0.625],
respectively. When all modalities are included, the performance gain is roughly equal to that of the best single
modality.

The differences between the merged and split literal configurations indicate that, despite our best efforts, information
from the node features has leaked into the structure. In the split configuration, the baseline performance is, as
expected, near random with an accuracy equalling that of a majority class classifier (0.495). However, in the merged
configuration the performance is roughly one-tenth higher than expected (0.616), indicating that some literals have an indegree
greater than one. Judging from the differences between modalities, these literals most likely express temporal or
visual information, which drop with roughly the same amount when moving from merged to split configuration.

\subsubsection{Link Prediction Results}

\begin{sidewaystable*}
\centering
	\caption{~Entity classification results for SYNTH in accuracy, averaged over 10 runs and with 95\% confidence interval,
	for both merged and split literals configuration. \textit{Structure} uses only the relational information whereas \textit{Structure + Features} also includes information from all supported modalities. The rest provides a breakdown per
    modality. All p-values are in relation to using only relational information.}
\label{tab:resultssynthclassification}
\smallskip
\begin{tabular}{lllcll}
\toprule
					  &\multicolumn{2}{c}{\textit{merged literals}}&&\multicolumn{2}{c}{\textit{split literals}}\\
					  \cmidrule(lr){2-3}\cmidrule(lr){5-6}
					  &\multicolumn{1}{c}{accuracy} & \multicolumn{1}{c}{p-value} && \multicolumn{1}{c}{accuracy} & \multicolumn{1}{c}{p-value} \\ \midrule
Majority Class        & 0.503 & - && 0.503 & -\\
Structure             & 0.616 ($\pm 0.003$) & - 			&& 0.497 ($\pm 0.005$) & -\\
Structure + Features  & 0.996 ($\pm 0.000$) & 2.33\e{-20} && 0.995 ($\pm 0.000$) & 5.09\e{-33} \\ \midrule 
Structure + Numerical & 0.744 ($\pm 0.011$) & 4.12\e{-03} && 0.785 ($\pm 0.012$) & 3.01\e{-29} \\
Structure + Temporal  & 0.763 ($\pm 0.019$) & 1.80\e{-14} && 0.625 ($\pm 0.012$) & 1.78\e{-14} \\
Structure + Textual   & 0.995 ($\pm 0.000$) & 2.39\e{-19} && 0.996 ($\pm 0.000$) & 3.97\e{-34} \\
Structure + Visual    & 0.642 ($\pm 0.063$) & 5.21\e{-04} && 0.556 ($\pm 0.044$) & 3.58\e{-53} \\
Structure + Spatial   & 0.957 ($\pm 0.002$) & 2.33\e{-20} && 0.949 ($\pm 0.001$) & 1.22\e{-30} \\ \bottomrule
\end{tabular}

\bigskip

\caption{~Link prediction results for SYNTH, averaged over 5 runs and with 95\% confidence interval,
		for both merged and split literals configuration. Listed are mean reciprocal rank (MRR) and hits@k with $k \in
		\{1,3,10\}$. \textit{Structure} uses only the relational information whereas
\textit{Structure + Features} also includes information from all supported modalities. The rest provides a breakdown per
modality. All p-values are in relation to using only relational information.}
\label{tab:resultssynthlinkprediction}
\centering\smallskip
\small
\begin{tabular}{llllllllllll}
\toprule
					  &\multicolumn{5}{c}{\textit{merged literals}}&&\multicolumn{5}{c}{\textit{split literals}}\\
					  \cmidrule(lr){2-6}\cmidrule(lr){8-12}
					  &\multicolumn{1}{c}{MRR} & \multicolumn{1}{c}{H@1} & \multicolumn{1}{c}{H@3} & \multicolumn{1}{c}{H@10} & \multicolumn{1}{c}{p-value} && \multicolumn{1}{c}{MRR} &\multicolumn{1}{c}{H@1} & \multicolumn{1}{c}{H@3} & \multicolumn{1}{c}{H@10} & \multicolumn{1}{c}{p-value} \\ \midrule
Structure             & 0.045 ($\pm 0.001$) & 0.041 & 0.048 & 0.050 & -&& 0.038 ($\pm 0.000$) & 0.032 & 0.045 & 0.046 & - \\
Structure + Features  & 0.069 ($\pm 0.009$) & 0.065 & 0.072 & 0.074 & 2.50\e{-05} && 0.057 ($\pm 0.003$) & 0.053 & 0.060 & 0.063 & 2.50\e{-05} \\
Structure + Numerical & 0.084 ($\pm 0.001$) & 0.081 & 0.085 & 0.088 & 2.50\e{-05} && 0.068 ($\pm 0.000$) & 0.064 & 0.071 & 0.075 & 2.50\e{-05} \\
Structure + Temporal  & 0.073 ($\pm 0.001$) & 0.070 & 0.074 & 0.078 & 2.50\e{-05} && 0.048 ($\pm 0.001$) & 0.043 & 0.056 & 0.060 & 2.50\e{-05} \\
Structure + Textual   & 0.030 ($\pm 0.003$) & 0.023 & 0.036 & 0.040 & 2.50\e{-05} && 0.035 ($\pm 0.000$) & 0.024 & 0.044 & 0.045 & 2.50\e{-05} \\
Structure + Visual    & 0.050 ($\pm 0.002$) & 0.044 & 0.053 & 0.063 & 2.50\e{-05} && 0.028 ($\pm 0.002$) & 0.026 & 0.029 & 0.034 & 2.50\e{-05} \\
Structure + Spatial   & 0.034 ($\pm 0.001$) & 0.028 & 0.038 & 0.041 & 2.50\e{-05} && 0.031 ($\pm 0.000$) & 0.022 & 0.040 & 0.041 & 2.50\e{-05}    \\ \bottomrule
\end{tabular}
\end{sidewaystable*}

\noindent Table~\ref{tab:resultssynthlinkprediction} reports the mean MRR and hits@$k$ over 5 runs on SYNTH, together
with its 95\% confidence interval and corresponding p-values. We use the same \textit{value\_merged}
[\textit{value\_split}] notation as before to express the performances in the merged and split configurations, respectively.

Overall, the results show that, for most modalities, including their information considerably improves the performance
when compared to the baseline (structure only). In all cases, these differences are statistically significant. When
information from all modalities is included, the performance also increases noticeably, irrespective of literal
configuration, from 0.045 [0.038] to 0.069 [0.057]. However, rather than performing roughly the same as the best performing
single modality (0.084 [0.068] for numerical information), including all modalities yields a performance that is slightly
lower. This contrasts with our classification results.

Similar to the classification results there is considerable variation between the performances per modality: including
just numerical information yields a large boost in performance, both for the merged and split literal configuration,
whereas including textual or spatial information results in a drop in performance to an MRR of 0.030 [0.035] and
0.034 [0.031], respectively. Also similar is the limited influence of including visual information, although a slight but 
significant gain to an MRR of 0.050 is still visible in the merged literal configuration.

A final observation is that there exists a difference in performance on the baseline of 0.007 between the split and
merged literal configurations, supporting our previous supposition that some information from the literals is encoded in
the graph's structure. As before, this effect seems most evident with temporal and visual information, both of which
drop considerably in performance from 0.073 to 0.048 and from 0.050 to 0.028, respectively, when changing from merged to split literals.

\subsubsection{Discussion}

\noindent Our results indicate that, in the most ideal setting, including node features in the learning process improves the
performance most or all of the times, depending on the task. This is most clear for node classification, which obtains a
significance performance boost irrespective of the modality we include. With link predication the results are less
clear cut, although most modalities seem to have a positive effect on the overall performance. However, since a perfect
score is practically unobtainable in this setting, it is difficult to gauge how much these effects actually matter or
whether we can achieve the same by simply running the baseline for a higher number of epoch. Similarly, the drop in
performance for some modalities might just as well be caused by the increased difficulty of the learning task. Some
support for this supposition might be found with the drop in performance when either textual or spatial information is
included, both of which require a relatively large number of parameters but still result in a near perfect score in
node classification. Another possible reason is that this dataset, which is optimized for classification, lacks 
properties that make it an ideal testbed for link prediction.

Despite the aforementioned differences between tasks, we would expect to see that each modalities affects the
performance roughly similar, especially with classification since literals from each modality carry a strong positive
signal. As our classification results show that this is not the case, any difference in performance in this task must
have originated in the MR-GCN and/or the dataset. For numerical and temporal information the precise cause is unclear
and more elaborate testing is needed to determine whether the less-than-perfect performance stems from our encoders, or
their implementation, or whether the fault lies with an imperfect data generation process. In contrast, since
we use the proven MobileNet architecture for our visual encoder, it is likely that our image generation process is to
blame for the lackluster performance when visual information in included.

When all modalities are included in the learning process, the overall performance approaches or equals that of the best
performing single modality. This suggests that the message-passing network largely succeeds in learning, by itself,
which information to include and which to ignore. This effect is again more profound in our classification results, for
which including all modalities yield near perfect accuracy, but is still visible in the link prediction setting. As
before, this difference between tasks may stem from the focus of the dataset on classification, resulting in less clear
signals when used for link prediction.


\subsection{Evaluation on Real-World Knowledge}
\label{sec:evalrw}

\noindent Whereas previously we evaluated the MR-GCN on synthetic knowledge, we here evaluate our implementation on real-world
knowledge graphs from various domains and with different (combinations of) modalities.

\subsubsection{Node Classification}

\noindent We evaluate the MR-GCN on five real-world knowledge graphs on node classification. General and
modality-specific statistics about each of these are listed in Table~\ref{tab:datasets} and~\ref{tab:datatypes},
respectively. A short description of each dataset is given next.

\begin{description}
	\item[AIFB+] The AIFB dataset is a benchmark knowledge graph about scientific publications from a research group,
		and about the people working there~\cite{ristoski2016collection}. This is the smallest of the datasets in our
		experiments, and lacks the datatype annotations needed to accurately determine the literals' modalities. These
		annotations were added by us, creating AIFB+.
	\item[MUTAG] MUTAG is a benchmark dataset about molecules, the atoms they consist of, and any mutagenic properties
		that they might have~\cite{ristoski2016collection}. This dataset only contains a single additional modality,
		encoded by numerical literals.
	\item[BGS] The BGS dataset contains information about geological measurements in Great Britain, and includes rock
		composition and	age~\cite{ristoski2016collection}. Also present is spatial information, in the form of point
		locations and polygons.
	\item[AM+] The Amsterdam Museum dataset (AM) is a benchmark knowledge graph which contains information about 
		the collection of a museum in The Netherlands~\cite{ristoski2016collection}. We use the AM+ version
		from~\cite{peter2020kgbench} in our experiments, which has been extended with datatype annotations and images,
		and which has a much higher number of labeled samples. 
	\item[DMG] The Dutch Monument Graph (DMG) is a benchmark dataset for multimodal entity
		classification~\cite{peter2020kgbench}. The DMG includes information from all five modalities listed in
		Section~\ref{sec:modalities} (in addition to relational information), with a strong emphasis on spatial information.
		The example given in Figure~\ref{fig:examplegraph} is from this dataset.

\end{description}

\begin{table*}[t]
\caption{Datasets used in our experiments. The AIFB+, MUTAG, and SYNTH datasets are used in both classification and link
	prediction, DMG, AM+, and BGS only for classification, and ML100k+ and YAGO3-10+ only for link prediction. 
	Literals with the same value are counted as the same node in the merged count, whereas they are counted separately
	in the split count.}
\label{tab:datasets}
\centering
\small
\begin{tabular}{llrrrrrrrr}
\toprule
Dataset            & & AIFB+  & MUTAG  & YAGO3-10+ & SYNTH    & ML100k+ & DMG     & BGS      & AM+        \\ \midrule
Relations          & & 46     & 24     & 44        & 42       & 13      & 60      & 104      & 33         \\
Entities           & & 2,835  & 22,540 & 50,639    & 16,386   & 56,204  & 148,127 & 103,055  & 1,026,150  \\
Literals		   & merged & 5,468  & 1,104  & 20,797    & 112,319  & 32,055  & 195,468 & 230,790  & 127,520    \\ 
                   & split & 8,705  & 11,185 & 32,448    & 132,790  & 115,495 & 488,745 & 386,254  & 799,660    \\ \midrule
Facts              & total & 29,219 & 74,567 & 167,848   & 181,942  & 227,399 & 777,124 & 916,345  & 2,521,035  \\
				   & train & 21,175 & 54,547 & 127,802 & 141,899 & 187,393 & - & - & - \\
				   & test  & 4,022  & 10,010 & 20,023 & 20,023 & 20,003 & - & - & - \\
				   & valid & 4,022  & 10,010 & 20,023 & 20,023 & 20,003 & - & - & - \\ \midrule
Classes            &       & 4      & 2      & -         & 2        & -       & 5       & 2        & 8          \\
Labeled            & total & 176    & 340    & -         & 16,384   & -       & 8,399   & 146      & 73,423     \\
                   & train & 112    & 218    & -         & 10,484   & -       & 5,394   & 94       & 33,423     \\
                   & test  & 36     & 68     & -         & 3,278    & -       & 2,001   & 29       & 20,000     \\
                   & valid & 28     & 54     & -         & 2,622    & -       & 1,001   & 23       & 20,000     \\ \bottomrule
\end{tabular}
\end{table*}

\begin{table*}[t]
\caption{Distribution of datatypes in the datasets. Numerical information includes all
subsets of real numbers, as well as booleans, whereas date, years, and other similar types are listed under
temporal information. Textual information includes strings and its subsets, as well as raw URIs (e.g.\ links). Images
and geometries are listed under visual and spatial information, respectively.}
\label{tab:datatypes}
\centering
\begin{tabular}{lrrrrrrrr}
\toprule
Dataset            & AIFB+ & MUTAG  & YAGO3-10+ & SYNTH  & ML100k+ & DMG     & BGS     & AM+    \\ \midrule
Numerical          & 115   & 11,185 & -         & 29,565 & 55,058  & 17,205  & 12,332  & 160,959 \\
Temporal           & 1,227 & -      & 12,447    & 44,207 & 55,661  & 1,800   & 13      & 202,304 \\
Textual            & 7,363 & -      & 10,001    & 29,540 & 3,200   & 398,938 & 279,940 & 376,150 \\
Visual             & -     & -      & 10,000    & 14,758 & 1,576   & 46,108  & -       & 58,855    \\
Spatial            & -     & -      & -         & 14,720 & -       & 20,866  & 73,870  & -      \\
Other              & -     & -      & -         & -      & -       & -       & 20,098  & -      \\ \bottomrule
\end{tabular}
\end{table*}

\paragraph{Results}

\noindent Table~\ref{tab:resultsmergedclassification} and \ref{tab:resultssplitclassification} list the results of our
classification experiments for merged and split literal configurations, respectively, and report the mean classification
accuracy over 10 runs on the test sets, together with its 95\% confidence interval. Corresponding p-values are available
in Appendix~\ref{sec:apx_results}. We once again use the \textit{value\_merged}
[\textit{value\_split}] notation to express the performances in the merged and split configurations, respectively.

Overall, our classification results show that the effects of including node features in the learning process are
considerable, influencing the performance both positively and negatively, and that these effects vary greatly between
datasets and modalities: including temporal information, for example, has a (slight) positive effect on the performance
on AIFB+, from an accuracy of 0.933 [0.883] to that of 0.939 [0.894], but including the same form of information with DMG
results in a noticeably performance drop from 0.717 [0.450] to 0.695 [0.400]. Similar effects
are observable for other modalities. Moreover, including all modalities does not necessarily result in a higher accuracy,
irrespective of dataset and literal configuration: only on AM+, do we observe an increase when learning on all
modalities, from an accuracy of 0.751 [0.578] to that of 0.760 [0.598]. 

Looking at the differences in baseline performance between the merged and split configurations, it is evident that all
datasets express some information from the literals in their structure. This is particularly clear in the case of DMG,
which drops considerably in performance from 0.717 to 0.450 when we keep literals with the same values as separate nodes. 
However, this effect does enable us to observe that including textual and spatial information significantly improves the
accuracy on DMG to 0.518 and 0.511, respectively. Similar on AM+ for textual information, which improves the performance
in the split literal configuration from 0.578 to 0.606. In both cases, the added value is masked when part of this
information is encoded in the structure. In contrast, the
baseline performance on BGS stays roughly the same (0.845 [0.849]), suggesting that only few literals share a value.

Finally, our tests indicate that only the results on DMG and AM+ are statistically significant. This is most likely the
result of the large number of labeled samples in the test sets of these datasets. Note that the difference of 0.001 on
DMG between the performance of the baseline and that of including all features in the split literal configuration is
still statistically significant because the Stuart-Maxwell test compares individual predictions rather than
accuracies.

\begin{table*}[t]
	\caption{Entity classification results in accuracy, averaged over 10 runs and with 95\% confidence interval, 
		with merged literal configuration. \textit{Structure} uses only the relation information whereas 
		\textit{Structure + Features} also includes information from all supported modalities. The rest provides a
	breakdown per modality. Corresponding p-values are reported in Table~\ref{tab:pvaluesmergedclassification}.
	Statistically significant results are annotated with $\dagger$. }
\label{tab:resultsmergedclassification}
\centering
\small
\begin{tabular}{llllll}
\toprule
Dataset               & AIFB+  			  & MUTAG  			  & DMG    		      & BGS              & AM+   		\\\toprule		   
Majority Class        & 0.415  			  & 0.621             & 0.478			  & 0.637             & 0.300  		\\		   
Structure             & 0.933 $(\pm 0.013)$ & 0.689 $(\pm 0.024)$ & 0.717 $(\pm 0.001)$ & 0.845 $(\pm 0.010)$ & 0.751 $(\pm 0.004)$ \\		
Structure + Features  & 0.908 $(\pm 0.011)$ & 0.658 $(\pm 0.001)$ & 0.475 $(\pm 0.028)^\dagger$ & 0.748 $(\pm 0.054)$ & 0.760 $(\pm 0.013)^\dagger$ \\\midrule
Structure + Numerical & 0.939 $(\pm 0.011)$ & 0.664 $(\pm 0.015)$ & 0.678 $(\pm 0.006)^\dagger$ & 0.828 $(\pm 0.000)$ & 0.756 $(\pm 0.006)^\dagger$ \\ 
Structure + Temporal  & 0.947 $(\pm 0.001)$ & -     			  & 0.695 $(\pm 0.001)^\dagger$ & 0.845 $(\pm 0.010)$ & 0.765 $(\pm 0.004)^\dagger$ \\ 
Structure + Textual   & 0.903 $(\pm 0.001)$ & -    			  & 0.538 $(\pm 0.012)^\dagger$ & 0.853 $(\pm 0.010)$ & 0.713 $(\pm 0.013)^\dagger$ \\ 
Structure + Visual    & -	  		      & -     			  & 0.466 $(\pm 0.028)^\dagger$ & -                 & 0.764 $(\pm 0.011)^\dagger$ \\ 
Structure + Spatial   & -	   			  & -     			  & 0.741 $(\pm 0.003)^\dagger$ & 0.807 $(\pm 0.045)$ & -       \\\bottomrule
\end{tabular}
\end{table*}

\begin{table*}[t]
	\caption{Entity classification results in accuracy, averaged over 10 runs and with 95\% confidence interval, 
		with split literal configuration. \textit{Structure} uses only the relation information whereas 
		\textit{Structure + Features} also includes information from all supported modalities. The rest provides a
		breakdown per modality. Corresponding p-values are reported in Table~\ref{tab:pvaluessplitclassification}.
	   Statistically significant results are annotated with $\dagger$.}
\label{tab:resultssplitclassification}
\centering
\small
\begin{tabular}{llllll}
\toprule
Dataset               & AIFB+  			  & MUTAG  			  & DMG    		      & BGS              & AM+   		\\\toprule		   
Majority Class        & 0.415  			  & 0.621  			  & 0.478			  & 0.637             & 0.300  		\\		   
Structure             & 0.883 $(\pm 0.017)$ & 0.662 $(\pm 0.000)$ & 0.450 $(\pm 0.004)$ & 0.849 $(\pm 0.010)$ & 0.578 $(\pm 0.004)$ \\		
Structure + Features  & 0.865 $(\pm 0.001)$ & 0.653 $(\pm 0.012)$ & 0.451 $(\pm 0.021)^\dagger$ & 0.829 $(\pm 0.019)$ & 0.598 $(\pm 0.018)^\dagger$ \\\midrule
Structure + Numerical & 0.869 $(\pm 0.011)$ & 0.655 $(\pm 0.004)$ & 0.369 $(\pm 0.011)^\dagger$ & 0.827 $(\pm 0.008)$ & 0.560 $(\pm 0.004)^\dagger$ \\ 
Structure + Temporal  & 0.894 $(\pm 0.001)$ & -     			  & 0.400 $(\pm 0.002)^\dagger$ & 0.841 $(\pm 0.010)$ & 0.515 $(\pm 0.005)^\dagger$ \\ 
Structure + Textual   & 0.861 $(\pm 0.011)$ & -    			  & 0.518 $(\pm 0.025)^\dagger$ & 0.852 $(\pm 0.010)$ & 0.606 $(\pm 0.012)^\dagger$ \\ 
Structure + Visual    & -	  		      & -     			  & 0.468 $(\pm 0.031)^\dagger$ & -                 & 0.594 $(\pm 0.004)^\dagger$ \\ 
Structure + Spatial   & -	   			  & -     			  & 0.511 $(\pm 0.003)^\dagger$ & 0.826 $(\pm 0.012)$ & -  	   \\ \bottomrule
\end{tabular}
\end{table*}

\subsubsection{Link Prediction}

\noindent We evaluate the MR-GCN for link prediction on four multimodal real-world datasets. Two of these---AIFB+ and MUTAG---
were also used in our node classification experiments, whereas the remaining two are exclusively used for link
prediction. The DMG and AM+ datasets are not used here, since their relative large number of facts would translate to
exorbitant long training durations. We also abstain from testing the MR-GCN on standard link prediction benchmark
datasets, such as FB15k-237 and WN18RR, as these lack node features. 

General and modality-specific statistics about each of the datasets are listed in Table~\ref{tab:datasets}
and~\ref{tab:datatypes}, respectively. All training, testing, and validation splits are stratified on predicate. A short
description of two datasets that are exclusively used for link prediction is given next. Because of the added complexity
accompanying link prediction, both datasets were subsampled to still allow for GPU acceleration.

\begin{description}
	\item[ML100k+] MovieLens-100k is a well-known benchmark dataset about users, movies, and ratings given to these movies
		by the users, and contains various information that includes, amongst others, the genders and ages of users, and the release
		dates and titles of movies~\cite{harper2015movielens}. We use a subset of the version introduced in~\cite{emnlp18}, which 
		extends the	original dataset with movie posters. This subset was generated by selecting the 500 users with the
		highest rating count, together with all information to which they are linked.

	\item[YAGO-10+] A popular link prediction benchmark dataset is the YAGO knowledge graph.
		Emphasizing general knowledge, the dataset contains various information about people, cities, countries, movies,
		and organizations~\cite{suchanek2007yago}. Similar as with ML100k+, we use a subset of the version introduced
		in~\cite{emnlp18}, which enriches the original graph with images, texts, and dates. The subset was generated by
		taking the intersection of all entities with images, texts, and dates, together with all information to which 
		they are linked.

\end{description}

\paragraph{Results} Table~\ref{tab:resultsmergedlinkprediction} and \ref{tab:resultssplitlinkprediction} reports the mean 
MRR and its 95\% confidence interval over 5 runs on the tests sets. Corresponding p-values and hits@$k$ statistics are
available in Appendix~\ref{sec:apx_results}. As before, we use the \textit{value\_merged}
[\textit{value\_split}] notation to express the performances in the merged and split configurations, respectively.

Overall, our results indicate that, for link prediction on real-world knowledge, including node features can have a
profound effect on the performance, and that this effect can be both positive and negative. For MUTAG, this effect
results in a considerable performance boost from an MRR of 0.162 [0.135] to that of 0.225 [0.202], whereas, for the
three remaining datasets, this effect results in a moderate drop in performance (e.g.\ AIFB+, from 0.252 [0.215] to
0.215 [0.161]) to a considerable drop (e.g.\ YAGO3-10+, from 0.053 [0.050] to 0.025 [0.021]). These results are
statistically significant for all datasets and configurations, except for AIFB+ which, when numerical information is
included, achieves roughly the same performance as the baseline.  A quick glance at Table~\ref{tab:datatypes} shows that
AIFB+ only contains few numerical literals, suggesting that this result is a poor indicator of the effect that including
numerical information has on the overall performance and can best be ignored. 

Similar to our classification results, there appears to exist no discernible pattern in the performances amongst
modalities.  Instead, here too, the results for individual modalities vary much between datasets. For MUTAG, for
example, adding numerical information results in a moderate performance boost from 0.162 [0.135] to 0.192 [0.140],
whereas, for ML100k+, including this form of information results in a decrease in performance from 0.124 [0.028] to
0.042 [0.004]. Also similar is that, when including information from all modalities, the overall performance seems to
roughly equal the average performance of all separate modalities combined.

The differences in baseline performance between the merged and split configurations shows that all datasets have some
information from the literals encoded in their structure. This is most evident for ML100k+, which drops from 0.124 to
0.028 when this information is lost. In contrast, the drop in performance on YAGO3-10+ is only minor ($\pm 0.003$),
indicating that only few literals have an indegree greater than one. Irrespective, for all datasets and configuration,
the performance in the split configuration is the same or worse than that in the merged setting.

\subsubsection{Discussion}

\noindent Our results on real-world knowledge show that, overall, the effects of including node features in the learning
process vary widely: for some datasets, including information from a certain modality results in a slight to considerable
performance boost, whereas for other datasets that same modality does little or even results in a performance drop. This
suggests that the potential gain of including node features strongly depends on the characteristics of the data and
on the strength of the signals provided by the modalities. Moreover, when all modalities are included, our results show
that the overall performance stays behind that of the best performing single modality. This could suggest that the
message-passing model has difficulties ignoring the negative signals, or that the positive signals lack sufficient
strength in many real-world datasets for the message-passing model to overcome this. 

Comparing the results on AIFB+ and MUTAG from our node classification and link prediction experiments shows that
the effect of including a modality on the performance differs between tasks. On AIFB+, for example, incorporating
temporal information results in a slight performance gain in the classification setting, whereas the opposite is true in
the link prediction setting. Similar on MUTAG for numerical information, which provides a considerable gain or drop in
performance depending on which problem we are trying to solve. These results suggest that the influence of certain
modalities on one task does not necessarily carry over to other tasks. A similar observation was made for our results on
artificial knowledge. However, since, here, none of the classification results on either dataset is statistically
significant, it remain unclear whether the differences between tasks really matter, or whether they stem from
instabilities caused by the small test sets.

\begin{table*}[t]
	\caption{Mean reciprocal rank (filtered), averaged over 5 runs and with 95\% confidence interval, 
		with merged literal configuration. \textit{Structure} uses only the relation information whereas 
		\textit{Structure + Features} also includes information from all supported modalities. The rest provides a
	breakdown per modality. Corresponding hits@$k$ and p-values are reported in Appendix~\ref{sec:apx_results}.
	Statistically significant results are annotated with $\dagger$.}
\label{tab:resultsmergedlinkprediction}
\centering
\begin{tabular}{llllll}
\toprule
Dataset               & AIFB+  			  & MUTAG  			  & YAGO3-10+	      & ML100k+           \\\toprule		   
Structure             & 0.252 $(\pm 0.006)$ & 0.162 $(\pm 0.008)$ & 0.053 $(\pm 0.002)$ & 0.124 $(\pm 0.014)$ \\		   
Structure + Features  & 0.215 $(\pm 0.004)^\dagger$ & 0.225 $(\pm 0.006)^\dagger$ & 0.025 $(\pm 0.001)^\dagger$ & 0.066 $(\pm 0.010)^\dagger$ \\\midrule
Structure + Numerical & 0.254 $(\pm 0.004)$ & 0.192 $(\pm 0.006)^\dagger$ & -                 & 0.042 $(\pm 0.006)^\dagger$ \\ 
Structure + Temporal  & 0.237 $(\pm 0.004)^\dagger$ & -     			  & 0.042 $(\pm 0.001)^\dagger$ & 0.111 $(\pm 0.012)^\dagger$ \\ 
Structure + Textual   & 0.213 $(\pm 0.005)^\dagger$ & -    			  & 0.021 $(\pm 0.002)^\dagger$ & 0.125 $(\pm 0.010)^\dagger$ \\ 
Structure + Visual    & -	  		      & -     			  & 0.024 $(\pm 0.001)^\dagger$ & 0.101 $(\pm 0.014)^\dagger$  \\ 
Structure + Spatial   & -	   			  & -     			  & -                 & - \\ \bottomrule
\end{tabular}
\end{table*}

\begin{table*}[t]
	\caption{Mean reciprocal rank (filtered), averaged over 5 runs and with 95\% confidence interval, 
		with split literal configuration. \textit{Structure} uses only the relation information whereas 
		\textit{Structure + Features} also includes information from all supported modalities. The rest provides a
	breakdown per modality. Corresponding hits@k and p-values are reported in Appendix~\ref{sec:apx_results}.
	Statistically significant results are annotated with $\dagger$.}
	\label{tab:resultssplitlinkprediction}
\centering
\begin{tabular}{llllll}
\toprule
Dataset               & AIFB+  			  & MUTAG  			  & YAGO3-10+	      & ML100k+           \\\toprule		   
Structure             & 0.215 $(\pm 0.004)$ & 0.135 $(\pm 0.009)$ & 0.050 $(\pm 0.001)$ & 0.028 $(\pm 0.008)$ \\		   
Structure + Features  & 0.161 $(\pm 0.003)^\dagger$ & 0.202 $(\pm 0.009)^\dagger$ & 0.021 $(\pm 0.001)^\dagger$ & 0.003 $(\pm 0.001)^\dagger$ \\\midrule
Structure + Numerical & 0.214 $(\pm 0.006)$ & 0.140 $(\pm 0.007)^\dagger$ & -                 & 0.004 $(\pm 0.001)^\dagger$ \\ 
Structure + Temporal  & 0.205 $(\pm 0.003)^\dagger$ & -     			  & 0.043 $(\pm 0.002)^\dagger$ & 0.002 $(\pm 0.000)^\dagger$ \\ 
Structure + Textual   & 0.154 $(\pm 0.006)^\dagger$ & -    			  & 0.022 $(\pm 0.003)^\dagger$ & 0.019 $(\pm 0.001)^\dagger$ \\ 
Structure + Visual    & -	  		      & -     			  & 0.022 $(\pm 0.002)^\dagger$ & 0.003 $(\pm 0.001)^\dagger$  \\ 
Structure + Spatial   & -	   			  & -     			  & -                 & - \\ \bottomrule
\end{tabular}
\end{table*}

\section{Discussion}

\noindent Our results show that including node features from various modalities can have a profound effect on the
overall performance of our models. However, the direction and magnitude of this effect differs depending on which dataset
we use, what modalities we include, and even which tasks we perform.

When learning on on artificial knowledge, our results indicate that including multimodal information can significantly
improve performance, and that the underlying message-passing model is capable of learning, by itself, which features to
including and which to ignore. This contrasts with our results on real-world knowledge, which show that including node
features can have very different effects depending on which dataset we use and what modalities we include. Moreover, the
same message-passing model seemed unable to overcome the negative influence of some of the modalities, sometimes even
resulting in an overall worse performance with node features than without. This difference between artificial and
real-world knowledge might have been caused by our decision to abstain from hyperparameter optimization. However, since
the same hyperparameters were effective on artificial knowledge, this is unlikely to produce such a large difference.
Similar for our choices of (neural) encoders, which were unchanged between experiments. Instead, it is more likely that
our chosen message-passing model has difficulties coping with negative signals and/or noise. This would
explain why weak, but still positive, signals such as the visual information in SYNTH pose no problem, whereas the
negative signals in some of the real-world datasets drag the overall performance down considerably.

A comparison of results between the merged and split literal configurations shows that the potential performance gain
from including node features is influenced by how much information from these features is already encoded in the
structure of a graph. In some cases, our results show that including the same information can have little effect in the
merged setting while providing a considerable performance boost in the split configuration. This suggests that much of
this information is already stored as relational information, and that we gain little by also feeding the raw values to
our model. This is not necessarily a problem if, by nevertheless including this information, the performance does
not decrease either. However, our results show that, for some datasets and modalities, including node features results in a
drop in performance. This might be caused by the added complexity that makes the problem more difficult to solve. 
Reducing the number of model parameters might be a first step to alleviate this problem (See also
Section~\ref{sec:futurework}).

Finally, we observed that only half the datasets used in our classification experiments---SYNTH, AM+, and DMG--- produced statistically
significant results. The datasets in question have a considerably higher number of labeled instances, allowing for a
more precise evaluation of the results. To accurately establish which model architectures performs well in this setting
we need more datasets with similarly sized test sets. However, the observed difference in statistical significance between
datasets with few and many labeled instances does suggest that the Stuart-Maxwell test is suitable to compare
classification results with. Similarly, in our link prediction experiments, we observed only a single result that lacked
statistical significance. A quick inspection suggested that this was justified, since the dataset---AIFB+---contained only
few features of the modality being tested. This suggests that the randomised paired t-test is suitable to validate link
prediction results with. Since most literature in this field forgoes with statistical testing, we hope that these results
encourage others to use these or similar tests for machine learning experiments on knowledge graphs.


\section{Conclusion}

\noindent In this work, we have proposed an end-to-end multimodal message passing model for multimodal knowledge
graphs. By embedding our model in the message passing framework, and by treating literals as first-class citizen, we
embrace the idea that this enables data scientists to learn end-to-end from any heterogeneous multimodal knowledge, as long
as it is represented as a knowledge graph. To test our hypothesis, we have implemented our model and evaluated its
performance for both node classification and link prediction on a large number of artificial and real-world knowledge
graphs from various domains and with different degrees of multimodality. 

Our results indicate that, overall, including information from other modalities can have a considerable effect on the
performance of our models, but that the direction and magnitude of this effect strongly depends on the characteristics
of the knowledge. In the most ideal situation, when the dataset contains little noise and strong positive signals,
incorporating node features has the potential to significantly improve performance. When faced with real-world
knowledge, however, our results show that this effect can vary considerable between datasets, modalities, and even
tasks.


Despite the mixed results on real-world knowledge, we believe that this work supports our hypothesis that by enabling our
models to naturally ingest literal values, and by treating these values according to their modalities, tailoring their
encodings to their specific characteristics, we stay much closer to the original and complete knowledge that is
available to us, potentially resulting in an increase in the overall performance of our models. 

Learning end-to-end on heterogeneous knowledge has a lot of promise which we have only scratched the surface of. A model
that learns in a purely data-driven way to use information from different modalities, and to integrate such information
along known relations, has the potential to allow practitioners a much greater degree of hands-free machine learning on
multimodal heterogeneous knowledge.


\subsection{Limitations and future work}
\label{sec:futurework}

\noindent Our aim has currently been to demonstrate that we can train a multimodal message passing model end-to-end which can
exploit the information contained in a graph's literals and naturally combine this with its relational counterpart,
rather than to established that our implementation reaches state-of-the-art performance, or even to measure its performance
relative to other published models. We therefore performed little hyperparameter tuning in our experiments, ensuring
that any observable difference in performance could be confidently attributed to the inclusion or exclusion of
information from a certain modality, rather than have been caused by a particular hyperparameter setting.

To properly establish which type of model architecture performs best in multimodal settings, and whether message passing
models provide an advantage over more shallow embedding models without message passing, we require more extensive,
high-quality, standard benchmark datasets with well-defined semantics (i.e.\ datatype and/or relation range
declarations) and a large number of labeled instances. Recently, some datasets have seen the light which are suitable
for this purpose (e.g.~\cite{peter2020kgbench}). However, to perform more precise evaluations and more accurate models comparisons,
we need even more datasets from a wide range of domains and with a large number of different modalities. Nevertheless, to determine
precisely what kind of knowledge is most fitting for this form of learning we are likely to require an iterative
process where each generation of models provides inspiration for the next generation of benchmark datasets and vice versa.

In other work, currently under submission, we explore techniques to reduce the overall complexity of a multimodal model
by reducing the number of parameters by merging some of the weight matrices. Our main motivation for this is the
necessity of full batch learning with many message passing networks---a known limitation---which makes it challenging to
learn from large graphs; a problem which becomes even more evident as we start adding multimodal node features.  Future
work will also investigate the other side of the spectrum by using a separate set of learnable weights per relation, as
opposed to sharing weights amongst literals of the same modality. While this adds some additional complexity, it allows
a more natural encoding of a graph in our model by capturing the semantics per relation. To illustrate this, compare
learning a single set of weights for age and height, both of which are numeric, against learning a separate set of
weights for each.

Lastly, a promising direction of research is the use of pretrained encoders. In our experiments, we show that the encoders
receive enough of a signal from the downstream network to learn a useful embedding, but this signal is complicated by
the message passing head of the network, and the limited amount of data. Using a modality-specific, pretrained encoder,
such as GPT-2 for language data~\cite{radford2019language} or Inception-v4 for image data~\cite{szegedy2017inception},
may provide us with good general-purpose feature at the start of training, which can then be fine-tuned to the specifics
of the domain.

\section*{Acknowledgments}

\noindent We express our gratitude to Lucas van Berkel for his insightful comments. This project is supported by the
NWO Startimpuls programme (VWData - 400.17.605) and by the Amsterdam Academic Alliance Data Science (AAA-DS) Program
Award to the UvA and VU Universities.

\begin{appendix}
	\section{Detailed Results}
	\label{sec:apx_results}

	\noindent The following tables list more detailed results from our experiments. 
	Tables~\ref{tab:pvaluesmergedclassification} and~\ref{tab:pvaluessplitclassification} list the statistical
	significance for our classification results for merged and split literal configurations, respectively.
	For our link prediction experiments,
	tables~\ref{tab:resultsaifblinkprediction},~\ref{tab:resultsmutaglinkprediction},~\ref{tab:resultsyagolinkprediction}, 
	and~\ref{tab:resultsml100klinkprediction} list the hits@$k$ and the statistical significance for AIFB+, MUTAG, YAGO3-10+,
	and ML100k+ respectively.

	\begin{sidewaystable*}
	\caption{~Statistical significance for entity classification results, averaged over 10 runs,
	for merged literal configuration. \textit{Structure} uses only the relational information whereas
	\textit{Structure + Features} also includes information from all supported modalities. The rest provides a breakdown per
	modality. All p-values are in relation to using only relational information.}
	\label{tab:pvaluesmergedclassification}
	\centering\smallskip
	\begin{tabular}{llllll}
	\toprule
						  &\multicolumn{1}{c}{AIFB+} & \multicolumn{1}{c}{MUTAG} & \multicolumn{1}{c}{DMG} & \multicolumn{1}{c}{BGS} & \multicolumn{1}{c}{AM+} \\ \midrule
	Structure + Features  & 5.72\e{-01} & 8.33\e{-02} & 1.96\e{-105} & 3.17\e{-01} & 5.12\e{-22}  \\ \midrule 
	Structure + Numerical & 1.00\e{-00} & 8.57\e{-02} & 6.62\e{-29}  & 1.00\e{-00} & 9.52\e{-03}  \\
	Structure + Temporal  & 1.00\e{-00} & -           & 3.21\e{-28}  & 1.00\e{-00} & 6.25\e{-112}  \\
	Structure + Textual   & 5.72\e{-01} & -           & 1.99\e{-108} & 3.17\e{-01} & 6.58\e{-87}  \\
	Structure + Visual    & -           & -           & 1.44\e{-49}  & -           & 7.91\e{-104}  \\
	Structure + Spatial   & -           & -           & 1.26\e{-02}  & 1.00\e{-00} & -  \\ \bottomrule
	\end{tabular}
	
	\bigskip
	
	\caption{~Statistical significance for entity classification results, averaged over 10 runs,
	for split literal configuration. \textit{Structure} uses only the relational information whereas
	\textit{Structure + Features} also includes information from all supported modalities. The rest provides a breakdown per
	modality. All p-values are in relation to using only relational information.}
	\label{tab:pvaluessplitclassification}
	\centering\smallskip
	\begin{tabular}{llllll}
	\toprule
						  &\multicolumn{1}{c}{AIFB+} & \multicolumn{1}{c}{MUTAG} & \multicolumn{1}{c}{DMG} & \multicolumn{1}{c}{BGS} & \multicolumn{1}{c}{AM+} \\ \midrule
	Structure + Features  & 6.40\e{-01} & 1.00\e{-00} & 2.26\e{-159} & 1.56\e{-01} & 8.26\e{-54}  \\ \midrule 
	Structure + Numerical & 1.00\e{-00} & 1.00\e{-00} & 6.14\e{-87}  & 3.17\e{-01} & 6.72\e{-48}  \\
	Structure + Temporal  & 8.01\e{-01} & -           & 3.82\e{-36}  & 3.17\e{-01} & 9.60\e{-135}  \\
	Structure + Textual   & 6.16\e{-01} & -           & 1.07\e{-144} & 1.00\e{-00} & 5.12\e{-58}  \\
	Structure + Visual    & -           & -           & 4.31\e{-89}  & -           & 1.40\e{-44}  \\
	Structure + Spatial   & -           & -           & 6.01\e{-18}  & 8.33\e{-02} & -  \\ \bottomrule
	\end{tabular}
	\end{sidewaystable*}

	\begin{sidewaystable*}
		\caption{~Hits@\textit{k} with $k \in
			\{1,3,10\}$ and statistical significance for link prediction results on AIFB+, averaged over 5 runs,
			for both merged and split literals configuration. \textit{Structure} uses only the relational information whereas
	\textit{Structure + Features} also includes information from all supported modalities. The rest provides a breakdown per
	modality. All p-values are in relation to using only relational information.}
	\label{tab:resultsaifblinkprediction}
	\centering\smallskip
	\begin{tabular}{llllllllll}
	\toprule
						  &\multicolumn{4}{c}{\textit{merged literals}}&&\multicolumn{4}{c}{\textit{split literals}}\\
						  \cmidrule(lr){2-5}\cmidrule(lr){6-10}
		& \multicolumn{1}{c}{h@1} & \multicolumn{1}{c}{h@3} & \multicolumn{1}{c}{h@10} & \multicolumn{1}{c}{p-value} && \multicolumn{1}{c}{h@1} & \multicolumn{1}{c}{h@3} & \multicolumn{1}{c}{h@10} & \multicolumn{1}{c}{p-value} \\ \midrule
	Structure             & 0.018 & 0.028 & 0.040 & -&& 	       0.015 & 0.024 & 0.034 & - \\
	Structure + Features  & 0.014 & 0.024 & 0.037 & 1.24\e{-04} && 0.010 & 0.017 & 0.027 & 1.24\e{-04} \\
	Structure + Numerical & 0.018 & 0.028 & 0.040 & 1.09\e{-01} && 0.015 & 0.024 & 0.034 & 1.13\e{-02} \\
	Structure + Temporal  & 0.016 & 0.027 & 0.036 & 2.60\e{-02} && 0.014 & 0.023 & 0.034 & 1.24\e{-04} \\
	Structure + Textual   & 0.014 & 0.023 & 0.036 & 1.24\e{-04} && 0.010 & 0.016 & 0.026 & 1.24\e{-04} \\
	Structure + Visual    & - & - & - & - && - & - & - & - \\
	Structure + Spatial   & - & - & - & - && - & - & - & - \\ \bottomrule
	\end{tabular}

	\bigskip

		\caption{~Hits@\textit{k} with $k \in
			\{1,3,10\}$ and statistical significance for link prediction results on MUTAG, averaged over 5 runs,
			for both merged and split literals configuration. \textit{Structure} uses only the relational information whereas
	\textit{Structure + Features} also includes information from all supported modalities. The rest provides a breakdown per
	modality. All p-values are in relation to using only relational information.}
	\label{tab:resultsmutaglinkprediction}
	\centering\smallskip
	\begin{tabular}{llllllllll}
	\toprule
						  &\multicolumn{4}{c}{\textit{merged literals}}&&\multicolumn{4}{c}{\textit{split literals}}\\
						  \cmidrule(lr){2-5}\cmidrule(lr){6-10}
		& \multicolumn{1}{c}{h@1} & \multicolumn{1}{c}{h@3} & \multicolumn{1}{c}{h@10} & \multicolumn{1}{c}{p-value} && \multicolumn{1}{c}{h@1} & \multicolumn{1}{c}{h@3} & \multicolumn{1}{c}{h@10} & \multicolumn{1}{c}{p-value} \\ \midrule
	Structure             & 0.014 & 0.017 & 0.026 & -&& 	        0.010 & 0.014 & 0.021 & - \\
	Structure + Features  & 0.016 & 0.025 & 0.035 & 4.99\e{-05} && 0.015 & 0.021 & 0.030 & 4.99\e{-05} \\
	Structure + Numerical & 0.014 & 0.021 & 0.029 & 4.99\e{-05} && 0.012 & 0.016 & 0.024 & 4.99\e{-05} \\
	Structure + Temporal  &  - & - & - & - && - & - & - & -  \\
	Structure + Textual   &  - & - & - & - && - & - & - & -  \\
	Structure + Visual    &  - & - & - & - && - & - & - & -  \\
	Structure + Spatial   &  - & - & - & - && - & - & - & -     \\ \bottomrule
	\end{tabular}
	\end{sidewaystable*}

	\begin{sidewaystable*}
		\caption{~Hits@\textit{k} with $k \in
			\{1,3,10\}$ and statistical significance for link prediction results on YAGO3-10+, averaged over 5 runs,
			for both merged and split literals configuration. \textit{Structure} uses only the relational information whereas
	\textit{Structure + Features} also includes information from all supported modalities. The rest provides a breakdown per
	modality. All p-values are in relation to using only relational information.}
	\label{tab:resultsyagolinkprediction}
	\centering\smallskip
	\begin{tabular}{llllllllll}
	\toprule
						  &\multicolumn{4}{c}{\textit{merged literals}}&&\multicolumn{4}{c}{\textit{split literals}}\\
						  \cmidrule(lr){2-5}\cmidrule(lr){6-10}
		& \multicolumn{1}{c}{h@1} & \multicolumn{1}{c}{h@3} & \multicolumn{1}{c}{h@10} & \multicolumn{1}{c}{p-value} && \multicolumn{1}{c}{h@1} & \multicolumn{1}{c}{h@3} & \multicolumn{1}{c}{h@10} & \multicolumn{1}{c}{p-value} \\ \midrule
	Structure             & 0.043 & 0.056 & 0.074 & -&& 	       0.044 & 0.052 & 0.064 & - \\
	Structure + Features  & 0.009 & 0.035 & 0.044 & 9.50\e{-04} && 0.005 & 0.030 & 0.038 & 9.50\e{-04} \\
	Structure + Numerical &  - & - & - & - && - & - & - & - \\ 
	Structure + Temporal  & 0.036 & 0.044 & 0.056 & 9.50\e{-04} && 0.036 & 0.045 & 0.056 & 9.50\e{-04} \\
	Structure + Textual   & 0.006 & 0.027 & 0.036 & 9.50\e{-04} && 0.008 & 0.028 & 0.039 & 9.50\e{-04} \\
	Structure + Visual    & 0.014 & 0.028 & 0.035 & 9.50\e{-04} && 0.011 & 0.028 & 0.036 & 9.50\e{-04} \\
	Structure + Spatial   &  - & - & - & - && - & - & - & -     \\ \bottomrule
	\end{tabular}

	\bigskip

		\caption{~Hits@\textit{k} with $k \in
			\{1,3,10\}$ and statistical significance for link prediction results on ML100k, averaged over 5 runs,
			for both merged and split literals configuration. \textit{Structure} uses only the relational information whereas
	\textit{Structure + Features} also includes information from all supported modalities. The rest provides a breakdown per
	modality. All p-values are in relation to using only relational information.}
	\label{tab:resultsml100klinkprediction}
	\centering\smallskip
	\begin{tabular}{llllllllll}
	\toprule
						  &\multicolumn{4}{c}{\textit{merged literals}}&&\multicolumn{4}{c}{\textit{split literals}}\\
						  \cmidrule(lr){2-5}\cmidrule(lr){6-10}
		& \multicolumn{1}{c}{h@1} & \multicolumn{1}{c}{h@3} & \multicolumn{1}{c}{h@10} & \multicolumn{1}{c}{p-value} && \multicolumn{1}{c}{h@1} & \multicolumn{1}{c}{h@3} & \multicolumn{1}{c}{h@10} & \multicolumn{1}{c}{p-value} \\ \midrule
	Structure             & 0.012 & 0.013 & 0.014 & -&& 	       0.022 & 0.030 & 0.039 & - \\
	Structure + Features  & 0.032 & 0.069 & 0.137 & 2.39\e{-05} && 0.001 & 0.003 & 0.006 & 2.39\e{-05} \\
	Structure + Numerical & 0.036 & 0.047 & 0.056 & 2.39\e{-05} && 0.002 & 0.004 & 0.007 & 2.39\e{-05} \\
	Structure + Temporal  & 0.090 & 0.119 & 0.146 & 2.39\e{-05} && 0.001 & 0.003 & 0.007 & 2.39\e{-05} \\
	Structure + Textual   & 0.114 & 0.127 & 0.138 & 2.39\e{-05} && 0.001 & 0.002 & 0.003 & 2.39\e{-05} \\
	Structure + Visual    & 0.092 & 0.103 & 0.120 & 2.39\e{-05} && 0.001 & 0.003 & 0.007 & 2.39\e{-05} \\
	Structure + Spatial   &  - & - & - & - && - & - & - & -  \\ \bottomrule
	\end{tabular}
	\end{sidewaystable*}

\end{appendix}

\bibliographystyle{plain}
\interlinepenalty=10000  
\bibliography{bibliography}

\end{document}